\documentclass[11pt,a4paper,dvipsnames]{article}
\usepackage[hyperref]{emnlp2021}
\usepackage{times}
\usepackage{latexsym}

\usepackage{float}
\usepackage[utf8]{inputenc}
\usepackage[T1]{fontenc}
\usepackage{tikz}
\usepackage{amssymb}
\usepackage{amsmath}
\usepackage{multirow}
\usepackage{array}
\usepackage{graphicx}
\usepackage{microtype}

\usepackage{algorithm2e} 

\usepackage{linguex} 

\title{Exploring the Promises of Transformer-Based LMs for the Representation of Normative Claims in the Legal Domain}

\author{Reto Gubelmann \\
  University of St.Gallen \\
  Rosenbergstrasse 30 \\
  9000 St.Gallen \\\And
  Peter Hongler\\
  University of St.Gallen\\
  Varnbüelstrasse 19 \\
  9000 St.Gallen \\
  \texttt{\{reto.gubelmann,peter.hongler,siegfried.handschuh\}@unisg.ch} \\\And
  Siegfried Handschuh \\
  University of St.Gallen\\
  Rosenbergstrasse 30 \\
  9000 St.Gallen \\ }
\date{}

\begin{document}
\maketitle

\begin{abstract}
In this article, we explore the potential of transformer-based language models (LMs) to correctly represent normative statements in the legal domain, taking tax law as our use case. In our experiment, we use a variety of LMs as bases for both word- and sentence-based clusterers that are then evaluated on a small, expert-compiled test-set, consisting of real-world samples from tax law research literature that can be clearly assigned to one of four normative theories. The results of the experiment show that clusterers based on sentence-BERT-embeddings deliver the most promising results. Based on this main experiment, we make first attempts at using the best performing models in a bootstrapping loop to build classifiers that map normative claims on one of these four normative theories. 

\end{abstract}

\tableofcontents

\section{Introduction}

Disagreements about normative claims are notoriously hard to resolve, and in some cases, they are even hard to recognize as such. For instance, consider \ref{ex:1}. Do you think that a tax system that follows this principle is just?

\ex. It is just to tax people with the same income equally.
\label{ex:1}

Example \ref{ex:1} illustrates what we mean by a normative claim: A moral judgment of some kind, that is, an assertion that something is either morally right or wrong. As we restrict our scope to tax law, the normative claims that we are interested in pertain to moral judgments of specific tax systems. Hence, while example \ref{ex:1} counts as a normative claim, example \ref{ex:2} does not count. While the latter is also about tax law, it does not make a claim about what is just or unjust in this domain, but rather what is legal.

\ex. It is illegal not to pay one's taxes.
\label{ex:2}

Furthermore, compare example \ref{ex:3}. This example is not normative in the same explicit sense as \ref{ex:1}: It does not directly make a claim about what is just; however, in contrast to \ref{ex:2}, its standing directly depends on an explicitly normative claim such as \ref{ex:1}. If one rejects the latter, one will reject \ref{ex:3} as well. We call statements of the kind of \ref{ex:1} \emph{directly normative}, while we call statements of the kind of \ref{ex:3} \emph{indirectly normative}.

\ex. We should attempt to create a tax system that taxes people with the same income equally.
\label{ex:3}

In the discussion on international tax law, claims of the kind of \ref{ex:1} are regularly made, and even more often they figure implicitly in the arguments of legal scholars, say when statements of the kind of \ref{ex:3} are used without further questioning or without further argument. However, very often, the authors do not make explicit the normative perspective from which they are arguing (for instance, by stating and defending a claim of type \ref{ex:1} with explicit recourse to the political-philosophical literature in this domain). This is not the result of ill will. It is becoming increasingly difficult for students as well as for practitioners in the field to keep an overview on the different normative positions in the field. However, without such an overview, the debate threatens to lose sight of the central normative presuppositions of their debates. In the worst case, adherents of different normative positions will retreat into their normative bubbles and hence permanently hinder any truly rational debate about these topics.

To move towards improving this situation, we explore the promises of using state-of-the-art LMs to cluster directly normative statements in tax law texts.  More specifically, we explore the capacities of various transformer-based LMs to cluster directly normative statements that have been identified by an expert as belonging to one of these four views together. We use a variety of configurations for our clusterers, including different clustering algorithms with a range of parameters, and we test clustering based on specific words as well as on whole sentences. Furthermore, we explore the promises of using the models that have shown to allow for the best clusterers to initiate a bootstrapping loop to build effective classifiers of indirectly normative statements. 

The task in focus of this article is both challenging and important. It is challenging because recognizing the normative background of a statement such as \ref{ex:1} requires expert knowledge, and even with such expert knowledge, genuine uncertainties remain in some cases. Furthermore, considered from a technical perspective, when compared to other clustering tasks, the amount of lexical overlap is substantially higher. If proponents of two different normative theories talks about tax justice, they mean something substantially different compared to a proponent of the procedural view. However, we are not confronted with clear-cut ambiguity, such as in the case of \emph{bank}. Both mean to capture the same idea, they just understand that idea quite differently.\footnote{In the philosophical and linguistic debate, such concepts are called ``essentially contested concepts''. The conception was first proposed by \cite{Gallie:1955}, for recent discussions see \cite{Collier:2006} and \cite{Rodriguez:2015}. According to this conception, concepts such as \textsc{Tax Justice} are such that essential parts of their meaning are disputed. And the reason for the dispute is that the disagreement is due to larger-scale differences in worldview.}

The task is important because the subject matter that is addressed in such normative arguments is of central importance for democratic societies. What counts as a just taxation system directly influences the set-up of a taxation system and impacts the lives of the members of that society. Hence, providing support to navigate such normative landscapes is of central importance for societies -- even more so if, as we have suggested, it is difficult for tax law researchers to be sensitive to these normative categories and the entire debate threatens to disintegrate into a number of normative bubbles.

After discussing related research, we introduce our datasets and experiment, we list the results and discuss them. We conclude by providing an outlook to further research.

\section{Related Research}

As mentioned above (section 1), we approach the clustering task in two different ways, one of them word-based, the other sentence-based. In the word-based case, our approach shares intriguing analogies to word-sense disambiguation (WSD), which is why we also introduce previous work in WSD. For the sentence-based approach, to the best of our knowledge, we cannot build upon more specific research than the generic and well-established clustering algorithms used for text and word clustering, which is why we focus on this research there. Furthermore, we also introduce the clustering algorithms with a focus on WSD, even though we will be using it to cluster not only words but also sentences -- the simple reason being that it still seems to be a related task to cluster words that have different meanings and statements of different senses of central legal concepts such as tax justice. 

\subsection{Word-Sense Disambiguation}

According to \citet[3]{Navigli:2009}, WSD is the task to ``computationally determine which sense of a word is activated by its use in a particular context''. Hence, in WSD, the goal is to decide, for any given word and context, which sense of the word in question is relevant for the context in question. Similarly, in our task of classifying normative statements, the goal is often to determine in what sense central concepts such as \emph{tax justice} are meant in a given context.

Word-Sense Disambiguation (WSD) has been labelled an AI complete problem, that is, a problem that is as difficult to solve as general artificial intelligence as a whole. This claim, usually attributed to the seminal survey by \citet{Navigli:2009}, but in fact first issued by \citet[47]{Mallery:1988}, is as catchy as it is hard to cash out. The claim is in analogy to the concept of NP completeness (nondeterministic polynomial) from complexity theory, which refers to problems that can currently not be solved directly, but only using numeric approximations. Despite this vagueness, the label is accurate insofar as WSD is among the most notoriously difficult tasks in NLP, with systems still struggling to beat very primitive baselines such as the most frequent sense baseline, where an ambiguous terms is simply always mapped onto its most frequent sense. We expect the task of mapping words or sentences onto normative categories to be similarly challenging as the mapping of words to distinct senses.

\paragraph{Supervised Approaches to WSD}

In supervised approaches to WSD, the system learns a classifier based on large amounts of sense-annotated texts. An intuitive example of such a supervised approach is k-Nearest Neighbor (kNN) classification. In this approach, the features of the labelled data are stored as individual data points in the feature space. Any occurrence of a word to be disambiguated is then matched against all of the examples stored; the one sense having most examples within the class of k nearest data points in the feature space is selected as the correct sense in the specific context.

While there is a general consensus in the field that supervised systems perform best, they do so only within domains where there is sufficient high-quality labelled training data. Hence, recent years have seen semi-supervised approaches receiving more attention. In semi-supervised approaches, researchers typically try to interpolate the existing labels to unlabeled samples, for instance by assigning the sample the one label whose features are closest (in cosine distance) to the unlabelled sample in question.

\citet{Papandrea:2017} present a toolkit for supervised word-sense disambiguation.  They have developed a Java API that represents the WSD tasks as consisting of four subtasks -- parsing, preprocessing, feature extraction, and classification. Researchers can replace the default elements that fulfill these subtasks with their own modules and train as well as test the resulting systems efficiently via the command line.

\citet{Duarte:2021} present an in-depth analysis of semi-supervised approaches. Their general framework consists of four steps: (1) they extract pos-tags and word embeddings both from the target word to be disambiguated and from a fixed number of context words and combine them all into one single feature vector. (2) They create a graph out of all the feature vectors using a kNN approach. (3) They propagate the labels to the unlabeled data. (4) They use the resulting model in disambiguation challenges. What makes the paper stand out is that, within this overall framework, they vary a large number of parameters including the label propagation algorithm used - they employ the popular local and global consistency \citet{Zhou:2004}, label propagation \citet{Yamaguchi:2016}, Gaussian fields \citet{Zhu:2003} and OMNI-prop \citep{Yamaguchi:2015}; they achieve the best results with label propagation. They also vary the embeddings used, with contextualized word embeddings from BERT \citep{Devlin:2019} and ELECTRA \citep{Clark:2020} performing best. Generally, the variation within a given benchmark is rather small, typically around 0.1 F1. \citet{Sousa:2020} use a very similar approach, also testing different combinations of label propagation algorithms with embeddings.

\paragraph{Knowledge-Based Approaches} In knowledge-based approaches to WSD, the occurrences of ambiguous words are disambiguated by relying on external knowledge. An early example of such a knowledge-based approach is the so-called \emph{Lesk} algorithm (see \cite{Lesk:1986}). The basic idea is to measure the word overlap between the immediate context of the occurrence and the different definitions of the senses. The algorithm then selects the one sense where the overlap is largest.

Knowledge-based approaches, while usually not reaching the precision of supervised ones, have the advantage that they can rely on existing knowledge-bases such as WordNet \citep{Miller:1995}. This makes knowledge-based approaches generally cheaper to extend to many different domains, as researchers do not have to create costly, time-consuming labelled datasets. 

On a very general level, most of the systems follow the same structure (see \cite{Scarlini:2020b}, \cite{Pasini:2020}, \cite{Bevilacqua:2020} for recent work that follows this structure). First, the words whose meanings are to be disambiguated are identified in a dataset, and the CWE of their occurrences in the corpus are extracted. For each word to be disambiguated, the extracted CWE are then clustered, say, using a k-means algorithm. Then, using a knowledge-base such as WordNet \citep{Miller:1995} and a matching algorithm such as the page-rank based UKB \citep{Agirre:2014}, the clusters are associated with the senses annotated in the knowledge base. In sum, this yields a dictionary of polysemous words, where each of the senses of each word is assigned a vector that characterizes the specific sense contextually. During runtime, the occurrence of the ambiguous words are then disambiguated by matching the CWE of the occurrence in question with the vectors stored and associated with each word: the occurrence of the word is mapped onto the one sense whose embedding is closest to the CWE in question.

\subsection{Clustering}

From the perspective of WSD, clustering approaches are a kind of unsupervised approaches. In their purest forms, these clustering approaches are no typical cases of word sense disambiguation, but rather a cases of word sense discrimination: typical clustering approaches that function without tagged training data yield clustered representations of occurrences that allow to identify semantically related uses of words, without (in the absence of explicitly listed senses, say from a knowledge base) mapping these clusters onto specific senses. 

A historical example of such an unsupervised word sense clustering algorithm is given by word spaces (see \cite{Schutze:1998} and again \cite[26-28]{Navigli:2009}). The underlying assumption of this approach is that ambiguities can be solved by looking at the co-occurrence patterns of ambiguous words. For instance, the biological sense of bug is supposed to co-occur with words that signal its biological constitution (food, habitat, reproduction, etc.), whereas the software sense is supposed to co-occur with IT vocabulary. Following this basic idea, each word is assigned a vector that represents its typical co-occurrences within a specified window (say, a number n words before and after the occurrence) in a given corpus, a so-called centroid, or context vector. The context vectors of an ambiguous word can then be clustered, say by agglomerative clustering, where, starting with singletons, iteratively further nearest members are added to the cluster until a threshold is reached.

A popular alternative to agglomerative clustering is k-means-clustering proposed by \citet{Lloyd:1982}, where the centroids are initialized randomly and then iteratively updated. See \cite[238]{Geron:2019} for an overview and a non-technical recipe. The basic algorithm runs as follows:

\begin{enumerate}
  \item Set the number of clusters $k$ as a hyperparameter
  \item Initialize the centroids randomly, say by randomly picking $k$ instances.
  \item Label the instances: Assign them to the closest centroid.
  \item Update the centroids
  \item \dots
\end{enumerate}

Over time, various improvements in efficiency to this simple algorithm have been made, such as trying to choose centroids that are far away from each other to initialize the algorithm \citep{Arthur:2006}, to avoid any unnecessary distance calculations \citep{Elkan:2003}, and to process large datasets in batches \citep{Sculley:2010}.

While k-means, especially when optimized as sketched, is very fast and scales well, it has its drawbacks. Notably, you need to specify the number of clusters $k$ as a hyperparameter, the algorithm performs by design poorly on datasets that have clusters of various sizes and non-spherical shape (it uses a simple distance metric, after all).

Another popular clustering algorithm is DBSCAN \citep{Ester:1996}. It's basic idea is to find regions of high density. One specifies as hyper-parameters the maximal distance $\epsilon$ and the minimal number of instances to be located within this distance. Then the algorithm finds core instances that have the minimal number of instances within $\epsilon$; if one of these instances in turn counts as a core instance by having at least the minimal number of instances within $\epsilon$, it is added to the cluster formed by the initial core instance, and so on.

Advantages of DBSCAN are that it copes well with clusters of very different shapes, and it has just two hyper-parameters -- no need to specify the number of clusters in advance, as with k-means. 

In our main experiment, we are using k-means and DBSCAN with a range of different parameters, in our exploration of two classifiers to initiate a bootstrapping loop, we are using kNN-classifiers. In the future, we plan to also explore the promises of label-propagation in the sense of unsupervised approaches to WSD as well as the creation of a domain-specific knowledge-base that could then be used for knowledge-based approaches.

\subsection{Transformer-Based LMs \& Classical Word Embeddings}

We use three different kind of model to deliver the embeddings for our experiments, which are all based on previous research: Transformer-Based LMs that deliver word-like embeddings, transformer-based LMs that have been fine-tuned with the specific goal to deliver high-quality sentence embeddings, and classical, non-transformer-based word embeddings. Since its publication in \citet{Vaswani:2017}, the transformer architecture has been very influential in virtually all domains of NLP, including natural language understanding (NLU). With the exception of two models, all of the models tested are derived from this basic architecture.

\paragraph{Word-based LMs}  We here use three well-researched transformer-based LMs, namely bert-base-cased and bert-large-cased \citep{Devlin:2019} as well as roberta-large \citep{Liu:2019},\footnote{These models were downloaded from huggingface.co, see \citet{Huggingface}.} 

\paragraph{Sentence-based LMs} We test a large number of SBERT-Models \citep{Reimers:2019}, as initial explorations showed that they perform clearly best. These SBERT-Models are based on a variety of transformer-based LMs (in addition to the classical BERT and RoBERTa, these are mpnet \cite{Song:2020}, distilroberta \cite{Sanh:2019}, xlm \cite{Lample:2019}, AlBERT \cite{Lan:2019}, and minilm \cite{Wang:2020}). 

\paragraph{Classical WE} The classical, static word embeddings, namely GloVE \citep{Pennington:2014} and Komninos \citep{Komninos:2016}, are included for purposes of comparison.\footnote{The sbert- as well as the classical models were optained from https://www.sbert.net/docs/pretrained\_models.html.}

\section{Datasets}

While we report the datasets for the main experiment and for the bootstrapping exploration separately, we here introduce the four normative categories that we have asked an expert to identify and that form the basis of both the clustering as well as the classifying experiment.

According to the so-called \emph{Deontological View}, a tax policy proposal is just if it focuses on the treatment of the tax payer and not on the distribution of the income within a society. Hence, according to the Deontological View, one should neither look at democratic procedures, nor at the effects that a given tax system would have on the economy. Rather, one should look at whether it conforms to basic moral principles, such as the equality of all human beings. In this sense, example \ref{ex:1} constitutes a clear instance of the Deontological View.

According to the \emph{Rawlsian View}, a tax system is just if it would be chosen by individuals that are under Rawls' famous veil of ignorance. Under this veil, individuals do not know their educational, financial, social, or any other position in the society whose tax system they are supposed to develop. It is generally agreed that such individuals would favor tax systems strongly focused on equality -- as they might end up as financially and educationally disadvantaged members of this society. For the purpose of the present analysis, a Rawlsian views emphasizes the need for redistribution from the rich to the poor. Of course, this is an oversimplification of Rawls theory. 

Taxation should mainly result from good, democratically grounded processes -- this is the gist of the \emph{Procedural View}. Such view includes positions that argue for a certain tax policy proposal based on a discussion or debate about the arguments against and in favor of such proposal.

The fourth and final normative theory used in this article is the \emph{Libertarian View}. According to it, taxation should be kept at a minimum in general, as it is considered illegitimate in all but a few cases. Obviously, this view strongly contrasts with the Rawlsian View, as the latter is much friendlier to redistribution of wealth, if it can be expected to contribute to equality.

\subsection{Main Experiment}

For the clustering, we asked the expert to manually choose 10 samples of each of the four normative categories identified, trying to find directly normative rather than indirectly normative statements -- where, of course, the boundary between the two is not always razor-sharp. This yields a total of 40 samples that were submitted to the clustering experiment. The samples are all grammatical sentences taken from publications in peer-reviewed journals from the legal domain, and their categorization was conducted by an expert. Then, a philosopher without special expertise in tax law went through the examples and annotated any disagreements. Where the disagreements could not be resolved by discussion, a different sample, whose categorization was uncontroversial, was chosen.

\subsection{Bootstrapping for Classifiers}

The input for our kNN-Classifying experiment is given by four articles focusing on tax law, belonging to various text sorts, where the expert suspected -- but did not antecedently identify -- indirectly normative claims. \textbf{Article 1} is a tax-related discussion directed at the educated public,\footnote{See here: \url{https://www.americanprogress.org/issues/economy/reports/2020/09/28/490816/capital-gains-tax-preference-ended-not-expanded/}, last consulted on 13 August 2021.} \textbf{article 2} is a research article, akin to the articles from which the 40 samples were taken.\footnote{We selected section B \citep[666ff.]{Kleinbard:2016} for our testing.} \textbf{Article 3} is a research article focusing on Chinese situations. This article has been published in a peer-reviewed journal focusing on tax issues in the Asia Pacific \citep{Xu:2021}. \textbf{Article 4} is a memo of a parliamentary debate from Canada,\footnote{https://sencanada.ca/Content/SEN/Committee/362/bank/rep/rep05may00-e.htm, last consulted on August 20, 2021.} Given the samples used to create classifiers, we expect the classifiers to return the best results on article 2, then on article 3, then on article 1, and finally on the parliamentary memo, that is, on article 4.

\section{Experiment: Clustering Directly Normative Statements}

For this clustering experiment, we took the 40 expert-chosen samples of directly normative statements and submitted it to a number of clustering algorithms. We tested different clustering algorithm (k-means vs. DBSCAN), each with a variety of parameters, different kinds of embeddings used for clustering, and a variety of pre-trained models to produce the embeddings. Details can be found on table \ref{tab:clusterconfig}.

\begin{table*}[h!] \centering \small 
\begin{tabular}{l|l|l|l} 
\hline  
  algorithms & algorithm-specific parameters & model & embedding-type\\
\hline
  \multirow{4}*{DBSCAN} & \multirow{4}*{min. members (3), eps (11)}& \multirow{2}*{word-based transformers (3)} & word-based\\ \cline{4-4} 
  && &mean of all words\\ \cline{3-4}
  && sentence-bert (17) & sentence-based\\ \cline{3-4} 
  && classical (2) & sentence-based\\
\hline
  \multirow{4}*{k-means} & \multirow{4}*{number of clusters (2)}& \multirow{2}*{word-based transformers (3)} & word-based\\ \cline{4-4} 
  && &mean of all words\\ \cline{3-4}
  && sentence-bert (17) & sentence-based\\ \cline{3-4} 
  && classical (2) & sentence-based\\
\end{tabular}
\caption{Overview on the configurations of the clustering systems tested. DBSCAN was used with three different minimal member counts, eleven different parameters for epsilon. k-means was tested with two different cluster sizes. Each of them was then tested with three different word-based LMs, which were each tested by focusing on a specific word as well as taking the mean of all words. Furthermore, we also tested 18 sentence-bert and two classical sentence-embedding models. Overall, this results in a total of 825 configurations for DBSCAN and 50 configurations for k-means tested. For details, see the appendix.}
\label{tab:clusterconfig}
\end{table*}

\paragraph{DBSCAN \& k-means: parameters} As introduced above, the two clustering algorithms tested here are among the most commonly used for clustering. For k-means, one has to specify in advance the number of clusters that should be learned. As the target number of clusters is 4, we wanted to give the algorithm some wiggle room incase the four categories are in fact better mapped onto five clusters (say, because the embeddings of some model are more naturally clustered into five clusters). For DBSCAN, two parameters have to be set (see above, section 2), namely the maximal distance between two members of the same cluster (called epsilon), and the minimal number of members of one single cluster. As it is more difficult, in our case, to guess good values for these parameters, we have tested the algorithm with a much wider range of parameters. For epsilon, the range is 2,2.5, ... 7, yielding 11 values, for the minimal number of members per cluster, the values tested are 2,3 and 4.

\paragraph{Models \& Embedding Types} We are testing three different kinds of models; for references, see above, section 2.3; for the full list of models, see thee appendix, table \ref{tab:models}. We use four different routines to extract the embeddings:

\begin{description}
  \item[Word-Based] In this routine, we use a list of pre-compiled words that are intended to encapsulate central concepts of dispute, such as tax, taxation, or VAT (for the full list of these words, see the appendix). In this version of the experiment, the clusterers are not clustering sentences, but rather these words, as they appear in the sentences. As there is more input from experts, we expected these clusterers to perform better than the others. Here, we use well-researched transformer-based LMs, namely RoBERTa and BERT (see above, section 2.3)
  \item[Sentence-Averaged Word-Based] In this routine, we use the average of all word embeddings, as the model delivers it for all words in the sentence. Hence, the sentence-embedding used here is the average of all word embeddings whose words appear in the sentence. Here, we use well-researched transformer-based LMs, namely RoBERTa and BERT (see above, section 2.3) 
  \item[Sentence-based] Here, we use the embeddings, as they directly result from the sentence-bert models trained by \citet{Reimers:2019}. These models also output the average of all word embeddings (which we manually compute in the second variant), but they have been fine-tuned on the sentence level by training them on a wide variety of sentence-level tasks and datasets (the original models reported in \cite{Reimers:2019} use the combination of the SNLI and the Multi-Genre NLI datasets). Furthermore, the models that they fine-tuning are of many flavors, ranging from classical BERT to recent proposals such as mpnet (see above, section 2.3).
  \item[Average of Classical Word Embeddings] We here test two classical kinds of word embeddings, GloVE as well as Kominos (see above, section 2.3), again taking the average of all word embeddings as the sentence embedding.
\end{description}

Overall, this resulted in 875 different clusterers. All of these clusterers were then given a simple task: to cluster the 40 sentences from the dataset.

\section{Results}

We report the results of the clustering by giving what we call the average weighed homogeneity (AWH) of a given clusterer. The weighed homogeneity expresses the ratio of the largest member category and the overall membership in the cluster, weighed by the relative size of the cluster. For instance, if cluster 0 has 10 members, 8 of which are Ralwsian, then the homogeneity of the cluster would be 8/10, whereas the weighed homogeneity would be 8/10 multiplied by 10/40 (as the overall number of samples is 40), resulting in a weighed homogeneity score of 0.2. We take the average of the weighed homogeneity of all clusters produced by the clusterer. Without weighing the homogeneity, a classifier could cheat by performing well in very small clusters, but very poorly in one large cluster where most of the samples are.

\begin{figure*}[h!]
  \includegraphics[scale=0.35]{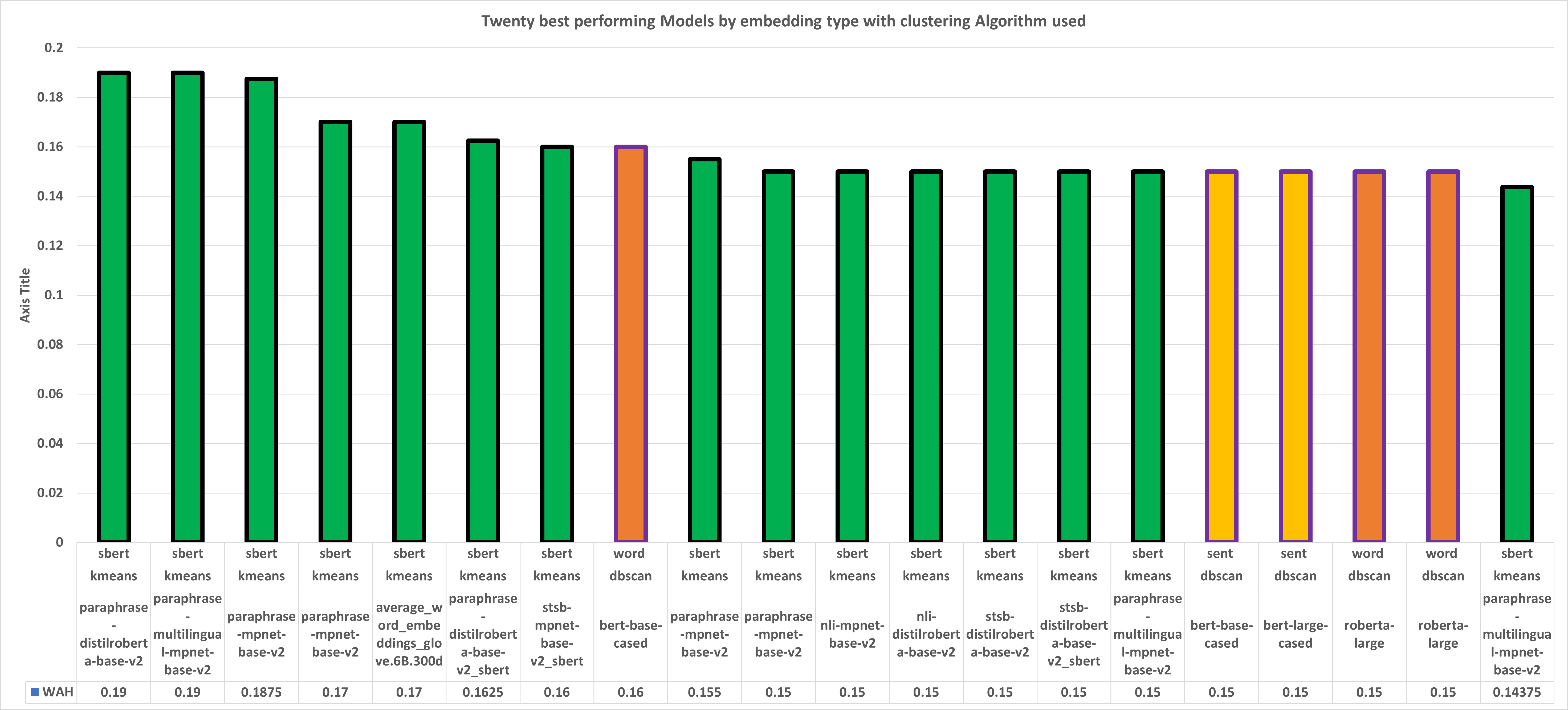}
\caption{Results of the first experiment. Only the twenty best performing clusterers are shown. green filling represents sbert-type embeddings, red word-based embeddings, yellow word-based averaged embeddings. Black framing of bars represents k-means clustering, purple framing represents DBSCAN clustering}
\label{fig:results_ex1}
\end{figure*}

Compare figure \ref{fig:results_ex1} for an overview on the results of this first experiment, showing the twenty best performing clusterers. For instance, the bar on the very left shows a score of 0.19, which represents the weighed average homogeneity that the model paraphrase-distilroberta-base-v2 could achieve using a k-means classifier (with specifications that are not shown on the chart, to facilitate overview). Overall, the chart shows that sbert-type embeddings perform clearly superior to both word and averaged-word embeddings. Surprisingly, embeddings not based on BERT, namely GLOVE-Average word embeddings land on rank 5 with a AWH-score of 0.17. 

\begin{figure}[h!]
  \includegraphics[scale=0.9]{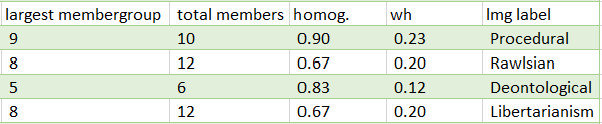}
\caption{Detailed clustering behavior of distilroberta.}
\label{fig:dist-best}
\end{figure}

\begin{figure}[h!]
  \includegraphics[scale=0.9]{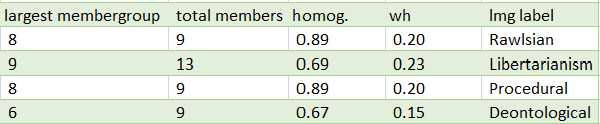}
\caption{Detailed clustering behavior of multilingual.}
\label{fig:mult-best}
\end{figure}

Figures \ref{fig:dist-best} and \ref{fig:mult-best} show the detailed clustering behavior of the two best performing clusterers. The former figure shows that the distilroberta-based embeddings allow for a surprisingly good clustering. Each of the four categories constitutes the largest membergroup in one of the four clusters (the same holds for the multilingual-based clusterer, see figure \ref{fig:mult-best}). For instance, the first cluster listed in figure \ref{fig:dist-best} contains 10 members in total, 9 of which belong to the procedural category. Furthermore, looking at figure \ref{fig:results_ex1}, it is clear that these two models are not very exceptional, but rather the best of a number of similarly well-performing clusterers.

\section{Discussion}
\subsection{Clustering}

The results of the clustering experiment surpassed our optimistic expectations. The two best models deliver embeddings that allow for clusterers which obviously latch onto the characteristics of the four categories. Even without any explicit instruction, the clusterers identified the four categories, in some cases, as in the first cluster of distilroberta, with impressive precision. 

Furthermore, experiment one shows that k-means outperforms DBSCAN in this task, a finding that was not affected by the very substantial variations in the parameters of DBSCAN used in the experiment. This is what one would expect given the data: As mentioned above (section 2.2), DBSCAN is good at finding areas of high densities. However, it seems that in this context, there are hardly any areas of high densities, and if there are, they do not latch onto the important categories. k-means, in contrast, is able to deal with distributions of various densities -- given the right hyperparameter, it always delivers the corresponding number of clusters, even if the differences in densities within each cluster is very large. 

Similarly, the sbert-models performed clearly best. Taking individual words or means over individual words, in contrast, performed poorer. On the one hand, this is plausible, as these models were developed specifically to deliver high-quality sentence embeddings. On the other hand, we expected the word-based approaches to perform well, too, as they were given further expert-compiled clues as to the specific words that are central for the task.

Finally, it is very interesting that a rather small model -- one based on distilroberta -- as well as a multilingual model outperform the large and monolingual models. This largely confirms the rankings on the sbert-page for clustering\footnote{https://www.sbert.net/docs/pretrained\_models.html, last consulted on August 13, 2021.}, while it does not answer the question as to why smaller models are of better use for clustering algorithms than larger ones. The multilingual model, finally, invites multilingual explorations.

In short, what the results of the first experiment show is that the basic set-up of this approach is very promising.

\subsection{Classification for Bootstrapping}

Encouraged by the results of our main experiment, we decided to explore the promises of using the insights gained by the experiment to build classifiers that could initiate a bootstrapping loop between experts and the classifiers. Hence, the question that we are trying to answer with this final exploration is whether classifiers built on merely 40 samples could be used to mine legal texts and yield useful suggestions for further normative statements that could then, after having been reviewed by experts, be used to build better classifiers, etc.

As a consequence of the results obtained in the clustering tests, we decided to focus on two sbert-models. Using these models to obtain the embeddings, we then trained two simple kNN-classifiers (k=3) that classify a given sentence as belonging to one of the four categories according to the three nearest neighbors of that sentence's embedding. To avoid classifying clearly non-normative sentences, we computed the centroid of the forty samples and empirically determined a threshold to automatically exclude any sentences that are beyond this threshold: Any embedding whose cosine similarity with the centroid is less than 0.6 is considered non-normative and not classified.

We then ran the dataset through the five articles mentioned and manually determined precision and recall. As this determination again requires expert knowledge in political philosophy, we asked an expert in the field of normative discussions in tax law as well as a philosopher to independently tag the results.

The results are shown in table \ref{tab:res_classifying}. We only counted as true positive a result if it also returned the correct categorization -- in addition to simply correctly realizing that a sentence was normative. 

\paragraph{Article 1} Overall, the classifier has split the article up in 55 sentence (or sentence-like) elements. Of these, 3 are normative in either the Rawlsian, Procedural, Deontological, or Libertarian sense in focus here. The distilroberta-based classifier has returned 3 positives, two of which are true positives. The multilingual-based classifier has returned four positives, three of which are true positives

\paragraph{Article 2} Overall, the classifier has identified 100 different sentences, 24 of which contain normative statements. The distilroberta-based classifier returns 14 positives, eight of which are true positives. The multilingual-based classifier returns 22 results, 11 of which are true positives.

\paragraph{Article 3} Overall, the classifier has split the article up in 99 sentences. Of these, 6 are normative in the specific ways in focus here. The multilingual-based classifier has 14 positives, 5 of which are true positives, and 9 are false positives. The distilroberta-based models shows only 2 positives, none of which are true positives.

\paragraph{Article 4} Overall, the classifiers have identified 100 sentences. Of these, 3 are normative in the specific ways in focus here. The distilroberta-based classifier has 2 positives, none of which are true positives. The multilingual-based classifier has 10 positives, two of which are true positives.

\begin{table}[h!]  \centering \small 
\begin{tabular}{l|l|l|l|l} 
\hline  
  Classifier & Art.1 & Art.2 & Art. 3 & Art. 4\\
\hline
distilroberta& 0.67/0.67 &0.57/0.3&0/0& 0/0\\
\hline
multilingual&0.75/1& 0.5/0.46&0.36/0.83& 0.2/0.67\\ 
\end{tabular}
\caption{Results of the classifying experiment. We report precision/recall.}
\label{tab:res_classifying}
\end{table}

The results of the two classifiers are encouraging, given the goal to build a classifier that can initiate a bootstrapping loop. In particular, it is notable that the two classifiers both deliver very few false positives, given the fact that in all four texts, normative claims were a small minority (roughly 10\% per document). These figures seem well enough to initiate the bootstrapping loop envisaged above. 

Furthermore, given the very simple calculation of the boundary between normative and non-normative, this is further evidence that the embeddings used in these experiments are promising for further analyses of normative judgments: Simple geometric properties of them can be used to draw a good distinction between normative and non-normative.

\section{Conclusion \& Outlook}

In this article, we have explored the promises of using well-known clustering and classifying approaches together with state-of-the-art transformer-based LMs to process normative statements in the legal domain. Our results indicate that this approach does indeed hold substantial promise. 

As our next steps, we plan the following:

\begin{enumerate}
  \item Using a bootstrapping loop, let experts and classifiers compile a large dataset for classifying. 
  \item Develop more sophisticated classifiers, systematically search the hyperparameter space.
  \item Use adversarial attack strategies to discern whether the classifiers are latching on merely to shallow lexical cues, or whether they are actually building on more sophisticated representations.
\end{enumerate}

\bibliographystyle{acl_natbib}

\bibliography{../Bibliography/habil}

\newpage
\appendix
\section{Details on Experiments}

Table \ref{tab:models} shows the full name of the models used.

\begin{table}[h!] \small
\begin{tabular}{|l|} 
\hline  
  Word-Based Models \\
  SBERT-Models \\ 
  Classical Models\\
 \hline
bert-base-cased \\
bert-large-cased \\
roberta-large \\
\hline
paraphrase-TinyBERT-L6-v2\\
paraphrase-distilroberta-base-v2\\ 
paraphrase-mpnet-base-v2\\ 
paraphrase-multilingual-mpnet-base-v2\\  
paraphrase-MiniLM-L12-v2\\ 
paraphrase-MiniLM-L6-v2\\ 
paraphrase-albert-small-v2\\ 
paraphrase-multilingual-MiniLM-L12-v2\\ 
paraphrase-MiniLM-L3-v2\\ 
nli-mpnet-base-v2\\ 
nli-roberta-base-v2\\ 
nli-distilroberta-base-v2\\ 
distiluse-base-multilingual-cased-v1\\ 
stsb-mpnet-base-v2\\ 
stsb-distilroberta-base-v2\\ 
distiluse-base-multilingual-cased-v2\\
stsb-roberta-base-v2 \\
\hline
average\_word\_embeddings\_glove.6B.300d \\
average\_word\_embeddings\_komninos \\
\hline
\end{tabular}
\caption{Overview on the models tested In clustering.}
\label{tab:models}
\end{table}

Following is a list of the words used for the word-based clustering routine, in this order (that is, words listed first take precedence):
\begin{enumerate}
\item taxation
\item tax*
\item VAT
\item revenue-raising
\item redistribution
\item income
\item reward
\item pay
\end{enumerate}

\end{document}